%% file: main.tex
\documentclass[letterpaper, 10 pt, conference]{ieeeconf}

\IEEEoverridecommandlockouts                              

\let\labelindent\relax
\usepackage{amsfonts}       

\usepackage{amsthm}
\usepackage{mathtools}      
\usepackage{amssymb}        
\usepackage{filecontents}
\usepackage{graphicx}       
\usepackage{marginnote}     
\usepackage{marvosym}       
\usepackage{overpic}        
\usepackage{tabularx}
\usepackage{cite}
\usepackage{color}
\usepackage[dvipsnames]{xcolor}
\usepackage[lined,linesnumbered,noend, ruled]{algorithm2e}
\usepackage{epstopdf}
\usepackage[inline]{enumitem}
\usepackage[normalem]{ulem}

\usepackage{lipsum}
\usepackage{subfig}     
\usepackage{wrapfig}    
\usepackage[font=footnotesize,labelfont=bf]{caption}    
\usepackage{microtype}  
\usepackage{comment}
\usepackage[group-separator={,}]{siunitx}   
\usepackage{titlesec}   

\usepackage{etoolbox}   
\makeatletter
\patchcmd{\@algocf@start}
  {-1.5em}
  {0pt}
  {}{}
\makeatother

\definecolor{darkblue}{rgb}{0.0, 0.0, 0.55}
\makeatletter
\let\NAT@parse\undefined
\makeatother
\usepackage[bookmarks=true, hidelinks]{hyperref}    
\hypersetup{
     colorlinks   = true,
     linkcolor= darkblue,
     citecolor    = darkblue,
     urlcolor=black
}

\theoremstyle{definition}

\theoremstyle{remark}

\SetKwProg{Fn}{Function}{}{}
\SetKwComment{Comment}{$\triangleright$\ }{}

\newcommand{\changed}[1]{\textcolor{black}{#1}}

\newcommand{\TODO}[1]{\textcolor{red}{#1}}

\DeclareMathOperator*{\argmax}{arg\,max}

\usepackage[utf8]{inputenc}
\usepackage[lined,linesnumbered,noend, ruled]{algorithm2e}

\title{\LARGE \bf
Parallel Monte Carlo Tree Search with Batched Rigid-body Simulations for Speeding up Long-Horizon Episodic Robot Planning
}
\author{Baichuan Huang\qquad Abdeslam Boularias\qquad Jingjin Yu     
\thanks{B. Huang, G. Teng, A. Boularias, and J. Yu are with the Department of 
Computer Science, Rutgers, the State University of New Jersey, USA. 
Emails: {\tt\small \{baichuan.huang, teng.guo, abdeslam.boularias, jingjin.yu\}@rutgers.edu}.
This work is supported by NSF awards 1734492, 1846043, 1845888, and 2132972.
}%
\vspace{-9mm}
}

\begin{document}

\maketitle

\begin{abstract} 
%
We propose a novel Parallel Monte Carlo tree search with Batched Simulations 
(PMBS) algorithm for accelerating long-horizon, episodic 
robotic planning tasks. 
Monte Carlo tree search (MCTS) is an effective heuristic search algorithm for 
solving episodic decision-making problems whose underlying search spaces are expansive. 
Leveraging a GPU-based large-scale simulator, PMBS introduces 
massive parallelism into MCTS for solving planning tasks through the batched 
execution of a large number of concurrent simulations, which 
allows for more efficient and accurate evaluations of the expected cost-to-go over large action spaces.
When applied to the challenging manipulation tasks of object retrieval from clutter, 
PMBS achieves a speedup of over $30\times$ with an improved  
solution quality, in comparison to a serial MCTS implementation. 
We show that PMBS can be directly applied to a real robot hardware with 
negligible sim-to-real differences. 
Supplementary material, including video, can be found at 
\href{https://github.com/arc-l/pmbs}{\texttt{\textcolor{blue}{https://github.com/arc-l/pmbs}}}. 
\end{abstract}

\section{Introduction}\label{sec:intro}
\input{texs/intro}

\section{Related Work}\label{sec:related}
\input{texs/related-works}

\section{Preliminaries}\label{sec:problem}
\input{texs/problem}

\section{Methodology}\label{sec:method}
\input{texs/method}

\section{Experimental Evaluation}\label{sec:experiments}
\input{texs/experiments}

\section{Conclusion}
\input{texs/conclusion}

\bibliographystyle{IEEETran}
\bibliography{references}


\end{document}

%% file: texs/intro.tex
The past decade has witnessed dramatic leaps in robot motion planning for solving problems  that involve
sophisticated interaction between the robot and its environment, with milestones 
including teaching quadrupeds to perform impressive tricks \cite{RoboImitationPeng20,hwangbo2019learning} 
and navigate challenging terrains \cite{KumarA-RSS-21}, enabling high-DOF robot hands to 
solve the Rubik's cube \cite{andrychowicz2020learning}, and so on. Whereas some of the
success can be attributed to the rapid advancement in deep learning \cite{krizhevsky2012imagenet} and deep reinforcement learning \cite{mnih2013playing},
another undeniable factor is the availability of fast high-fidelity 
physics engines, including PyBullet \cite{coumans2021} and MoJuCo \cite{todorov2012mujoco}.
These physics engines allow the simulation of the physics of complex rigid-body systems, 
sometimes faster than real-time, which enables the collection of large amounts of 
realistic system behavior data without even touching the actual robot hardware. 
Nevertheless, most physics simulators are CPU-based, which can only simulate a limited 
number of robots simultaneously; this has led to some studies seeking parallelism by 
using a massive amount of computing resources. For example, the OpenAI hand study 
\cite{andrychowicz2020learning} used a total of $6,144$ CPU cores to train their model 
for over $40$ hours, which is costly and time-consuming.

As physics simulation starts to become a bottleneck in solving robotic tasks, GPU-based physics engines have recently begun to emerge, including Isaac Gym 
\cite{makoviychuk2021isaac} and Brax \cite{brax2021github}, to address the issue by enabling large-scale rigid body simulation. Early results are 
fairly promising; for example, the training of the OpenAI hand using Isaac Gym
can be done on a single GPU in one hour, translating to a combined resource-time saving 
of several magnitudes. Similar success has also been realized in applying reinforcement 
learning on quadrupeds, robotic arms, and so on \cite{makoviychuk2021isaac}. 
\begin{figure}[t]
\vspace{1.5mm}
    \centering
    \includegraphics[width = 0.98\linewidth]{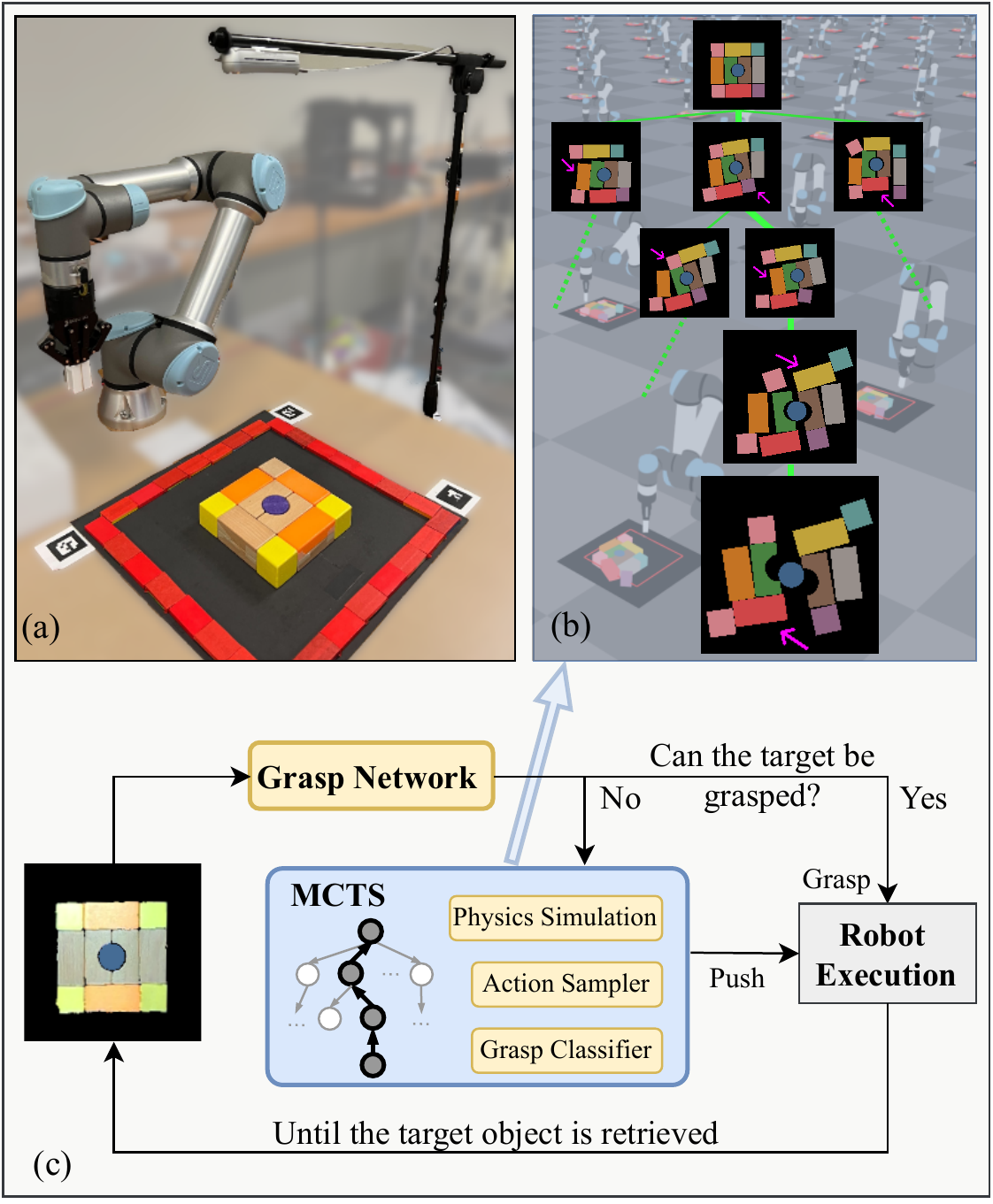}\label{fig:overview}
    \caption{(a) The hardware setup includes a Universal Robots UR-5e with a Robotiq 2F-85 two-finger gripper and an Intel RealSense D455 RGB-D camera. (b) Planning and simulation carried in physics simulator where thousands of virtual robots operate in parallel. (c) Overview of our system; the small blue cylinder at the center is the target object to be retrieved.
    \label{fig:overview}
    }
    \vspace{-8mm}
\end{figure}

In this work, we exploit the power of large-scale rigid body simulation for optimally 
solving long-horizon episodic robotic planning tasks, such as multi-step object retrieval 
from clutter, leveraging the strength of another powerful tool that has attracted a 
great deal of attention -- Monte Carlo tree search (MCTS) \cite{coulom2006efficient}. 
MCTS demonstrates clear advantages in solving long-horizon optimization problems 
without the need of significant domain knowledge \cite{silver2017mastering}, and was already employed for solving challenging manipulation tasks \cite{huang2021visual,huang2022interleaving}. However, 
even with significant guidance using domain knowledge \cite{huang2022interleaving}, MCTS incurs fairly 
long planning times due to its need of carrying out numerous rounds of
sequential \emph{selection-expansion-simulation-backpropagation} cycles. 
The long planning time, sometimes several minutes per decision step, limits the
applicability of the methods from \cite{huang2021visual,huang2022interleaving} toward real-time decision making. 

Through combining MCTS and large-scale rigid-body simulation with Isaac Gym
\cite{makoviychuk2021isaac}, and carefully introducing parallelism into the mix, we
have developed a new line of parallel MCTS algorithms for efficiently solving 
long-horizon episodic robotic planning tasks. 
The development of the large-scale rigid-body simulation enabled parallel MCTS is 
the key contribution of this research, which is highly non-trivial. This is 
because MCTS has an inherently serial characteristic; as will be explained in more 
detail, the \emph{selection} phase of an MCTS iteration depends on the 
completion of the previous selection-expansion-simulation-backpropagation iteration. 
Fusing MCTS and Isaac Gym for solving long-horizon manipulation planning tasks also 
brings significant technical integration challenges because many computational 
bottlenecks must be addressed for the parallel MCTS implementation to be efficient.

We call our algorithm \textbf{P}arallel \textbf{M}onte Carlo tree search with 
\textbf{B}atched \textbf{S}imulation (PMBS). 
As its name suggests, PMBS realizes parallel MCTS computation through batched rigid-body simulation enabled by Isaac Gym.
Efficiently combining MCTS and Isaac Gym, PMBS achieves over $30\times$ speedups in planning efficiency for solving the task of object retrieval 
in clutter, while still achieving better solution quality, as compared to an optimized serial MCTS implementation, using identical computing hardware.
PMBS drops the single-step decision making time to a few seconds on average, which is close to 
being able to solve the task in real-time. 
We further demonstrate that PMBS can be directly applied to real robot hardware 
with negligible sim-to-real differences.

%% file: texs/related-works.tex
\noindent\textbf{Task and Motion Planning}. 
Our investigation of long-horizon episodic planning tasks falls 
under the general umbrella of \emph{task and motion planning} (TAMP) \cite{kaelbling2011hierarchical,srivastava2014combined,toussaint2015logic,dantam2016incremental}. A characteristic that 
differentiates TAMP and other episodic tasks, e.g., playing 
rule-based games \cite{silver2017mastering}, is the inherently
uncountable infinite decision space induced by physical
interactions (e.g., pushing, grasping, and so on).
The vast search space suggests that the injection of domain 
knowledge is likely necessary to enable effective planning, 
e.g., using a combination of symbolic reasoning and
sampling-based motion planning techniques \cite{garrett2020pddlstream,migimatsu2020object}. 
More recently, data-driven methods have also seen increased 
application to solving TAMP, e.g., using learning to guide 
the search process \cite{chitnis2016guided, kim2019learning},
predicting the outcome of actions \cite{driess2020deep,huang2021dipn}, directly predicting feasibility \cite{wells2019learning}, combining neuro-symbolic task planning and motion primitives \cite{zhu2021hierarchical}, and so on. 

\noindent {\bf Object Retrieval.} 
Object retrieval from clutter, the long-horizon task that we focus on in this study, can be viewed as a form of rearrangement planning \cite{DBLP:journals/corr/abs-1912-07024,GaoFenYuRSS21}.
%
%
%
Online planning for object search with partial observations has been discussed in~\cite{8793494}.
Retrieving objects under occlusion was also recently considered in~\cite{DBLP:journals/corr/abs-1903-01588} where parallel-jaw and suction grasping were used along with pushing to de-clutter surroundings of target objects. 
%
%
%
A model-free reinforcement learning technique has also been used for searching for objects in~\cite{DBLP:journals/corr/abs-1911-07482}.
In~\cite{kurenkov2020visuomotor},  an agent was trained to find a continuous trajectory of a gripper that pushes away clutter or pushes the target object to free space, mimicking human-like behavior.
%
%
A human in-the-loop solution was proposed in~\cite{DBLP:journals/corr/abs-1904-03748} to help with searching for objects in clutter.
%
%
A deep Q-Learning method~\cite{xu2021efficient} considers a similar task and setup but uses additional primitives such as sliding objects from the top.
%
%
%
%
%
Our work partially builds on~\cite{huang2021visual,huang2022interleaving}, which used MCTS for object retrieval, but with the goal of significantly accelerating the planning process. 
%

\noindent {\bf Grasping and Singulation.}
The retrieval task that we tackle is closely related to 
other manipulation tasks including grasping and singulation.
Challenges including friction modeling and inertia estimation
has led to the arise of data-driven grasping methods~\cite{DBLP:conf/iros/BoulariasKP11,lu2020multifingered}.
Recently, grasping in clutter has received more attention~\cite{DBLP:conf/aaai/BoulariasBS14,BoulariasBS15, kalashnikov2018qtopt, wen2021catgrasp}.
Convolutional Neural Networks are widely used to construct grasp proposal networks such as Dex-net 4.0~\cite{mahler2019learning}, which are trained to detect 6D grasp poses in point clouds~\cite{DBLP:journals/corr/PasGSP17}.
%
%
%
Singulation, i.e., isolating specific object(s) from the rest~\cite{6224575}, is necessary for object retrieval.
%
%
Usually, a sequence of pushing and grasping actions is used to clear the clutter that surrounds the target object.
In~\cite{10.1007/978-3-030-28619-4_32}, a \emph{model-free} method was used to learn a reactive pushing policy without long-horizon reasoning.
Later, other model-free reinforcement learning algorithms~\cite{zeng2018learning, 8560406} used learned push policies to improve grasping.
%
%
%

%% file: texs/problem.tex
\subsection{Problem Formulation}
In this paper, we task a robot equipped with a camera and a two-finger gripper to grasp a desired object from a densely packed clutter, as a concrete instance of 
long-horizon episodic robot planning problems. 
The workspace is a confined planar surface.
Two types of primitive actions are allowed: pushing and grasping.
All objects are rigid; the target object has a different color to facilitate its detection.
The only observation available to the robot is an RGB-D image that is taken by a top-down fixed camera, as shown in Fig.~\ref{fig:overview}. 
Every time the robot executes a push or a grasp action, a new image is taken.
A similar problem has been previously defined in~\cite{huang2021dipn, huang2021visual, huang2022interleaving}.
Compared to~\cite{huang2021visual, huang2022interleaving}, the problem addressed in the present work is significantly more challenging to solve because the workspace is confined to a substantially smaller area, while keeping the number and sizes of objects the same. Consequently, the free space between the objects is reduced, and the robot needs to find a larger number of shorter surgical push actions in order to free the target object and grasp it. In fact, we found from our experiments (Sec.~\ref{sec:experiments}) that the original setup considered in~\cite{huang2021visual, huang2022interleaving} can be solved using a brute-force parallel search in a GPU-based physics simulator, without a Monte Carlo tree search. 


The {\bf object-retrieval-from-clutter} task can be formalized as a Markov Decision Process (MDP) defined as $(\mathcal{S, A, T, R, O, \gamma})$, to minimize the number of pushes needed for retrieving the target. $\gamma\in[0,1]$ is a discount factor.
This MDP is described in the following.  

{\bf State space $(\mathcal{S})$} is a bounded workspace $W$ containing $n$ objects. A state $s_t$ at time $t$ is defined as $(robot_t, obj_t^1, obj_t^2, \dots, obj_t^n)$ wherein
$robot_t$ is a vector of the robot's joint angles and  
$obj_t^i$ is a vector of the object $i$'s pose and geometry. 
\noindent {\bf Action space $(\mathcal{A})$} is the union of two subsets: push actions $A^p$ and grasp actions $A^g$.
A push $a^p \in A^p$ is a quasi-static planar motion defined as $a^p=(x_s, y_s, x_e, y_e)$ where $(x_s, y_s)$ and $(x_e, y_e)$ are the start and end points of the gripper tip in a straight line motion.
A grasping action $a^g \in A^g$ is a parallel jaw grasp defined as $a^g = (x, y, z, \theta)$, where $(x,y,z)$ is the center point between the gripper's two fingers, and $\theta$ is the rotation angle in the $z$-axis of the workspace. The opening distance of two fingers is fixed.
\noindent {\bf Transition function $(\mathcal{T})$} maps a state $s_t$ to $s_{t+1}$ given action $a_t$ according to laws of quasi-static motion and friction, while tacking into account collisions. 
\noindent {\bf Reward function $(\mathcal{R})$} returns a scalar $r=\{0, 1\}$ given input state $s_t$. If a target object is grasped, then $r_t=1$, otherwise $r_t=0$. All push actions are given a zero immediate reward. \noindent {\bf Observation space $(\mathcal{O})$} contains RGB-D images $o_t$ taken from the top-down camera.


\subsection{Monte Carlo Tree Search}
Monte Carlo tree search (MCTS) \cite{coulom2006efficient} was notably employed for playing turn-based games like chess and go. As a result, MCTS is clearly applicable to long-horizon episodic planning tasks. 
MCTS builds a search tree, balancing exploration and exploitation, by iteratively performing \emph{selection-expansion-simulation-backpropagation} operations. In the \emph{selection} phase, MCTS selects a best node to grow the tree. A popular node selection criterion is based on the \emph{upper confidence bound} (UCB) \cite{auer2002finite,kocsis2006bandit}, 
\begin{equation}\label{eq:ucb1}
    \argmax\limits_{n' \in \text{children of } n} \frac{Q(n')}{N(n')} + c \sqrt{\frac{2\ln{N(n)}}{N(n')}}, 
\end{equation}
where $Q(n)$ is the sum of rewards collected starting from the state corresponding to node $n$, $N(n)$ is the number of times $n$ was selected so far. The selection process continues until it finds a node that corresponds to a terminal state or a node that has never-explored children.
We note that a node $n$ is always associated with a state $s$ and an observation $o$; sometimes a node $n$ and the corresponding state $s$ are used interchangeably. 

After a node $n$ is selected, if it is not a terminal node, it will be \emph{expanded} and its new child, say $n'$, will be added to the search tree. Subsequently, a \emph{simulation} will be carried out at $n'$. This \emph{selection-expansion-simulation} process is repeated until a terminal state (or a stopping condition) is reached, which yields a reward. 
The obtained terminal reward is \emph{propagated back} from $n'$ all the way to the root node, while updating the sum of reward ($Q(n)$) and incrementing the number of visits ($N(n)$) for all the nodes along the path. 

The entire selection-expansion-simulation-backpropagation procedure is repeated for several rounds, after which MCTS selects the action at the root node that yields the child with the best average reward to execute on the real robot. The operations of MCTS are visualized in Fig.~\ref{fig:mcts} (the upper box). 

\begin{figure}[ht!]
\vspace{1.5mm}
    \centering
    \includegraphics[width = 0.95\linewidth]{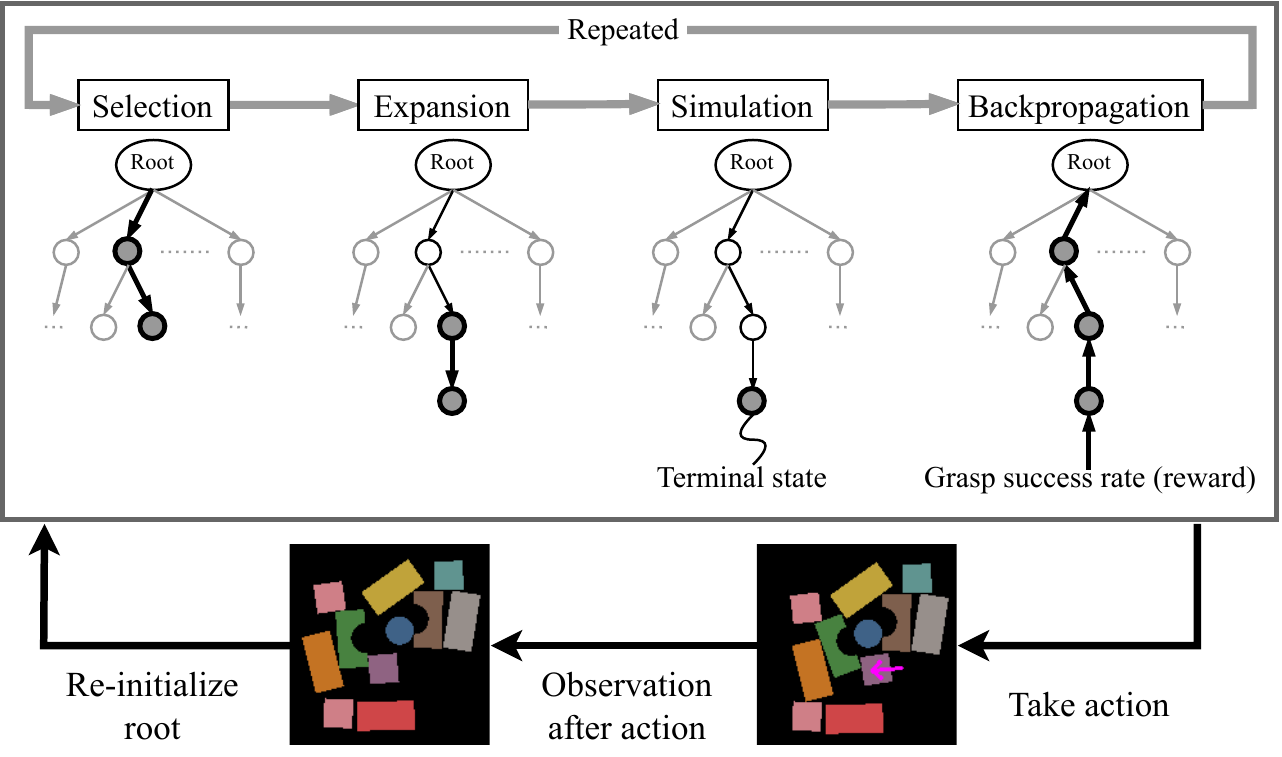}\label{fig:mcts}
    \caption{Outline of Monte Carlo tree search as adapted for object retrieval.
    \label{fig:mcts}
    }
    \vspace{-7mm}
\end{figure}

%% file: texs/method.tex
Effectively employing MCTS to tackle long-horizon episodic robot planning requires a highly non-trivial adaptation of MCTS. In this section, we first describe the necessary preparation for integrating MCTS and physics simulation for object retrieval, then augmentations to the architecture for GPU-based processing, and then outline our key ideas and design choices in our parallelization effort. 

\subsection{Serial MCTS for Object Retrieval from Clutter}\label{subsec:smcts}
%
To use MCTS for the object retrieval task and solve real instances, we integrate it in a process that alternates between search in simulation and execution on the real system. 
Our MCTS process takes in a scene that is segmented into objects, and uses physics simulation to reason about the proper push actions to facilitate the final retrieval of the target object. %
An overview of the MCTS process is provided in Fig.~\ref{fig:overview}.
In other words, we first replicate in the simulator the real perceived scene at the beginning of each episode, perform computation and simulation, and then execute with the real robot the action that results from the simulation to guide the resolution of the retrieval task on the real objects. 

We now describe the details of our basic serial MCTS adaptation. For the selection step, the standard UCB formula, Eq.~\eqref{eq:ucb1}, is used. For the expansion step, for a selected node $n$ that has not been expanded, we sample many potential push directions by examining the contour of the objects. These sampled pushes become the candidate actions under $n$ for expansion. 
After a sampled push action is chosen, the action is executed in the physics simulation and a new node is added to the tree. The MCTS simulation step is then carried out with additional consecutive random pushes to obtain a reward for the newly added node. Note that for each simulation step, we must decide whether the resulting state is a terminal state; this is done using a \emph{grasp classifier}, to be explained later. 

An important design decision we make here, to render MCTS computation more tractable, is to limit the depth of the tree.
We limit the depth of the overall tree to be no more than some $d_T$.
The simulation can be carried out for at least $d_s$ steps.
This means that the maximum depth reached by MCTS does not exceed $d_T + d_s$.
If expansion happens at depth $d_T$, we allow the state to be simulated further until $d_T + d_s$.
Given our goal of finding the least number of pushes for retrieving the target object, $d_T$ and $d_s$ can potentially be dynamically updated when an identified terminal node has a depth $d$ smaller than $d_T$. In this case, we set $d_T = d$ and $d_s = 0$. We terminate an MCTS process if: (1) the elapsed time exceeds a preset budget $T_{\text{max}}$, (2) the tree is fully explored, or (3) the target can be grasped in an explored node and all nodes at its parent's level have been explored.

After each full MCTS run, we execute the best action it returns on the actual scene (simulated or real), and then use a \emph{grasp network} (GN)  \cite{huang2021visual,huang2022interleaving} to tell us whether the target object is retrievable. If it is, GN further tells us how to grasp it; the task is then completed. Details of GN, for replication purposes, can be found in the online supplementary material.

\subsection{Adaptions for GPU}\label{subsec:gpua}
Besides simulation, which can be sped up using GPU-based physic engines, there are three additional bottlenecks in the process to parallelize MCTS for object retrieval. One of these these is the parallelization of MCTS itself and the other two are specific to the object retrieval problem: action sampling and grasp feasibility prediction. We leave the first bottleneck for Sec.~\ref{subsec:pmbs} and address the latter two here. 

\begin{wrapfigure}[8]{r}{1in}
\vspace{-4.5mm}
  \includegraphics[width=1in]{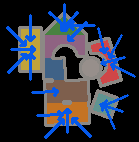}
  \vspace{-5mm}
  \caption{\label{fig:action}
    Sampled push actions.}
\end{wrapfigure}
\noindent \textbf{Speeding up Action Sampling}. 
Because the number of push action choices is uncountably infinite, action sampling is necessary. 
%
\changed{
We modified the action sampler from \cite{huang2021visual, huang2022interleaving} with slight changes and a more efficient implementation.
}
As shown in Fig.~\ref{fig:action}, for a given $o_t$, actions $a^p_t$ are sampled around the clutter.
\changed{
$N_a$ actions are evenly sampled around the contour of each object, from edge to center.
}
Actions that cannot be executed due to collisions are discarded.
Further speedups are obtained by pre-computing the sampled actions for each object and only performing collision checking \changed{between the robot's start pose of push and objects} at runtime.


\begin{wrapfigure}[11]{r}{1.4in}
\vspace{-5mm}
  \includegraphics[width=1.4in]{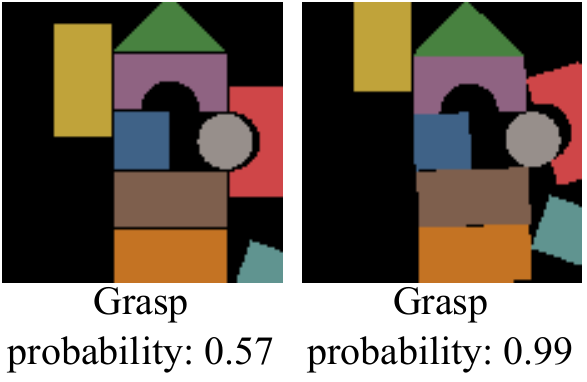}
  \vspace{-6mm}
  \caption{\label{fig:grasp}
    Examples of using the \emph{grasp classifier} to produce probabilities to grasp the object at center (\textcolor{RoyalBlue}{blue} in this case). Here we used an RGB image for illustration purpose (input should be a depth image).}
\end{wrapfigure}
\noindent\textbf{Grasp Evaluation}.
Previous learning-based methods for object retrieval use a \emph{grasp network} (GN) to evaluate the feasibility of grasping the target object \cite{huang2021visual, huang2022interleaving}, which becomes a time-consuming bottleneck when parallelized directly.
%
GN is relatively slow because it evaluates a large number of possible grasp poses. However, knowing the best grasp pose is unnecessary if we only want to know how ``graspable'' a state is. 
Given this observation, we develop a simplified \emph{grasp classifier} (GC) that only returns a grasp probability (Fig.~\ref{fig:grasp}). 
GC is an EfficientNet-b0~\cite{tan2019efficientnet} that takes a depth image as input and outputs a logit between 0 and 1.
Given a depth image and a target object, we can query GC whether the target object is graspable by comparing its output to a preset threshold $R_c^{*}$. Details about GC's implementation and training in simulation can be found in the online supplementary material. Note that GN is still used after each full MCTS run for potentially grasping the target object, as described in Sec.~\ref{subsec:smcts}.
\subsection{Parallel MCTS with Batched Simulation}\label{subsec:pmbs}
Given the availability of powerful GPU-based physics simulators including Isaac Gym 
\cite{makoviychuk2021isaac} and Brax \cite{brax2021github}, which enables the simulation of a large number of systems independently and simultaneously, a natural route for speeding up long-horizon episodic robot planning tasks is to introduce parallelism into the MCTS pipeline outlined in Sec.~\ref{subsec:smcts}, to perform many simultaneous simulations. 
However, it is challenging to introduce parallelism into MCTS because optimal node selection depends on the reward of all previous rounds.
%
To enable parallelism in MCTS for object retrieval from clutter and harness GPU-based simulation, we introduce the following modifications to the MCTS procedure outlined in Sec.~\ref{subsec:smcts}.
We assume the number of parallel environments in the simulator is fixed to some number $N_e$. Each parallel environment contains an identical virtual robot and objects.

\noindent\textbf{Selection with Virtual Loss.}
By observing the operations of MCTS, it is not difficult to see that the parallelism of MCTS requires modifying the UCB formula. Otherwise, the same node in a search tree will be selected for expansion in multiple parallel environments, leading to redundancy and poor performance. 
%
To address this issue, we borrow the idea of \emph{virtual loss}~\cite{chaslot2008parallel}, which has shown to give good results in multiple application domains~\cite{Liu2020Watch, yang2021practical, silver2018general}. 
Virtual loss is used to adjust the calculation of UCB values for the nodes that have been selected but not yet expanded~\cite{chaslot2008parallel, mcts2012},
\begin{equation}\label{eq:ucb2}
    \argmax\limits_{n' \in \text{children of } n} \frac{Q(n')}{N(n') + \hat{N}(n')} + c \sqrt{\frac{2\ln{(N(n) + \hat{N}(n))}}{N(n') +\hat{N}(n')}},
\end{equation}
where the $\hat{N}(n)$ is the number of selected but not yet expanded nodes under node $n$.
$\hat{N}(n)$ will be reset to zero once the selection phase of parallel MCTS is completed.
Basically, Eq.~\eqref{eq:ucb2} penalizes selecting nodes that have already been selected in some other parallel environment but for which expansion and simulation have not yet been completed. 
With Eq.~\eqref{eq:ucb2}, it is still possible for a node $n$ to be selected multiple times, which may lead to redundant simulations. 
%
To avoid this and ensure that no redundant simulations are carried out, we mark all selected actions of node $n$ and share this information across all the parallel environments. 

A collection of state-action pairs is returned from the selection phase. 
The same state could be selected many times, but all state-action pairs in the selected collection are unique.
For example, in Fig.~\ref{fig:parallel-mcts}, upper left, $(s_{t+2}^1, a^1), \ldots, (s_{t+1}^4, a^4)$ are four such state-action pairs.
Batch-mode expansion/simulation on this collection is then performed in parallel using GPU.

\begin{figure}[ht!]
\vspace{2mm}    
    \centering
    \includegraphics[width = 0.8\linewidth]{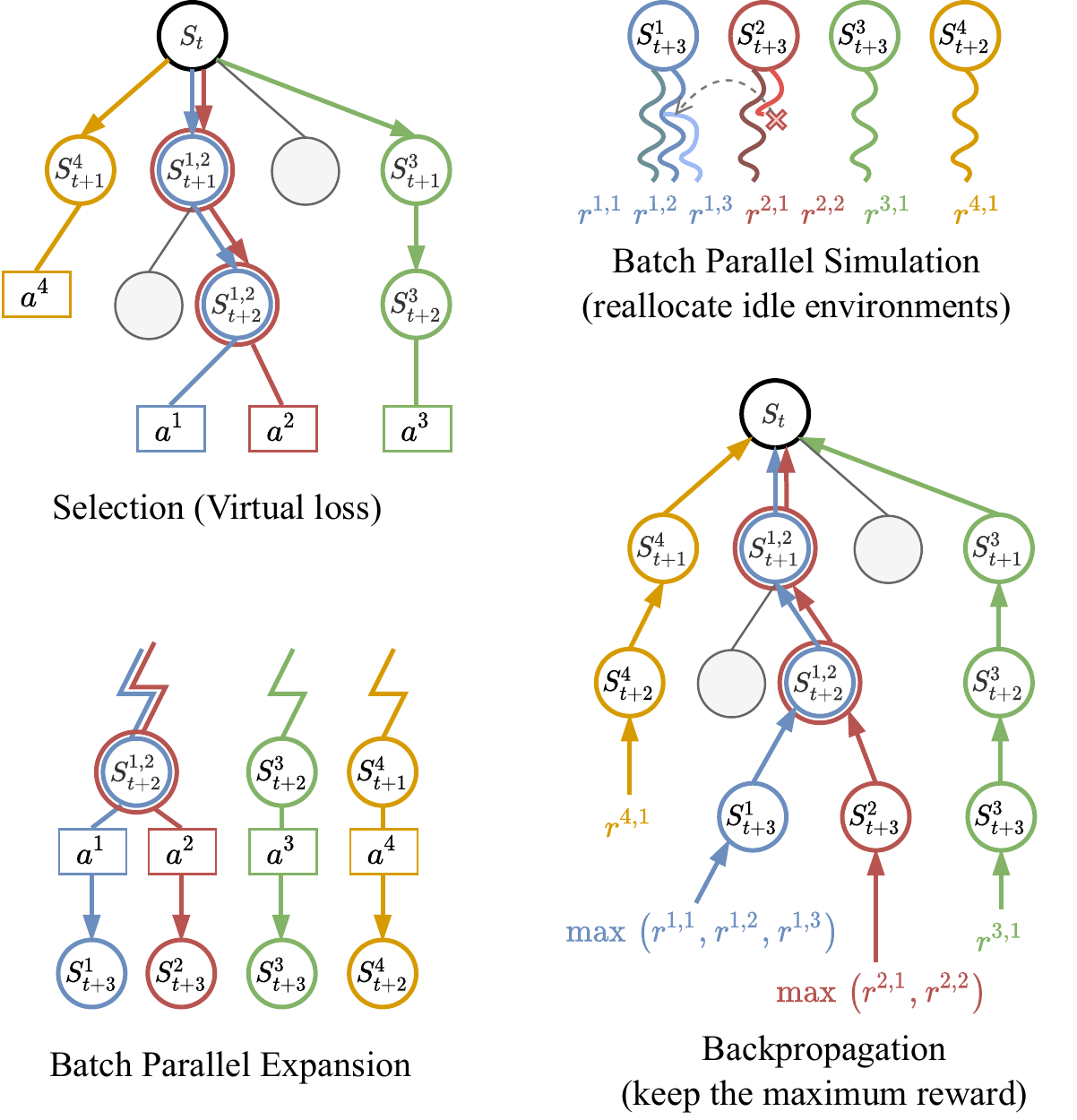}\label{fig:parallel-mcts}
    \caption{Steps in PMBS, our parallel MCTS with batch operation.\label{fig:parallel-mcts}
    }
\vspace{-3mm}    
\end{figure}

\noindent \textbf{Batch Expansion.}
After a batch of state-action pairs ($\{(s, a)\}$) has been selected, the expansion step is carried out for all of these pairs simultaneously. For this purpose, environments in the simulator (Isaac Gym) will be loaded with the appropriately selected states, after which expansion (transition) is carried out in parallel. A batch expansion creates a set of new nodes added to the tree, each of which is different. 
In Fig.~\ref{fig:parallel-mcts}, lower left, $s_{t+3}^1, \ldots, s_{t+2}^4$ are the result of expanding $(s_{t+2}^1, a^1), \ldots, (s_{t+1}^4, a^4)$, respectively.
%
\begin{algorithm}[h]
    \begin{small}
    \DontPrintSemicolon
    \SetKwFunction{FMain}{Main}
    \SetKwFunction{FMCTS}{Parallel-MCTS}
    \SetKwFunction{FSelect}{Selection}
    \SetKwFunction{FExpand}{Expansion}
    \SetKwFunction{FSimulation}{Simulation}
    \SetKwFunction{FBackpropagation}{Backpropagation}
    \Fn{\FMain{$s_t, o_t$}}{
        \While{\normalfont there is a target object in workspace}{
            \If{\normalfont the target object can be grasped (query GN)}{Execute grasp the target object}
            \lElse{Execute \FMCTS{$s_t$}\tcp*[f]{Push}}
        }
    }
    \vspace{1mm}
    \Fn{\FMCTS{$s$}}{
        Create root node $n_0$ with state $s$ \;
        $\text{es\_level} \gets 1$ \tcp*[f]{Early stop level}\;
        $\text{graspable\_nodes} \gets \varnothing$ \;
        \While {\normalfont (within time budget) and (depths of all graspable\_nodes are greater than es\_level)}{
            $[(n^1,a^1), \dots, (n^{N_\text{e}},a^{N_\text{e}})] \gets \FSelect(n_0)$  \; 
            
            Reset all $\hat{N}(n)$ to 0 \;
            
            $[{n'}^1, \dots, {n'}^{N_\text{e}}] \gets \FExpand([(n^1,a^1), \dots, (n^{N_\text{e}},a^{N_\text{e}})])$ \;
            
            \For{\normalfont$n'$ in $[{n'}^1, \dots, {n'}^{N_\text{e}}]$}{
                \If{\normalfont $GC(n'(o)) > R_c^{*}$}{
                    $\text{graspable\_nodes} \gets \text{graspable\_nodes} \cup \{n'\}$ 
                }
            }
            
            \If{\normalfont all nodes at $\text{es\_level} - 1$ are fully expanded or terminal}{
                $\text{es\_level} \gets \text{es\_level} + 1$
            }
            
            $[r^1, \dots, r^{N_\text{e}}] \gets \FSimulation([n_1', \dots, n_{N_\text{e}}'])$\;
            
            \FBackpropagation{$[({n'}^1, r^1), \dots, ({n'}^{N_\text{e}},r^{N_\text{e}})]$}
            
            \Return the $a^p$ that leads to best child node of root, ranked by Eq.~\ref{eq:ucb2}
        }
    }
    
    \caption{\label{alg:pmcts} Parallel MCTS with Batched Simulation}
    \end{small}
\end{algorithm}

\noindent \textbf{Batch Simulation.}
The batch simulation step of our parallel MCTS implementation is similar to the batch expansion step, but with additional steps inserted before and after. Before a push simulation, random actions must first be selected, using the action sampling method outlined in Sec.~\ref{subsec:gpua}. After each push simulation, GC, as described in Sec.~\ref{subsec:gpua} is applied to evaluate the outcome.
As we can see, to best exploit the parallelism from the simulator, action sampling and GC should be carried out as efficiently as possible, so that they do not become significant computational bottlenecks. 

During simulation, we also perform \emph{leaf parallelization} \cite{chaslot2008parallel} when the number of simulation environments is more than the number of states for which MCTS simulations are to be carried out. This is reflected in Fig.~\ref{fig:parallel-mcts}, upper right, where the first two states each are simulated twice initially. 
If some environments, after a push, are predicted by GC as graspable, then further simulation on these environments will not be carried out, and these environments can be re-purposed. For example, in Fig.~\ref{fig:parallel-mcts}, upper right, a simulation under the second state terminates early, and the associated environment can be used to perform additional simulation for the first state.

\noindent \textbf{Backpropagation.}
The backpropagation phase is straightforward to execute, as it simply backpropagates the rewards to the root of the tree. 
We note that, for a single state for which multiple simulations are carried out, it is natural to select the maximum reward obtained instead of taking averages (see Fig.~\ref{fig:parallel-mcts}, lower right).

The pseudo-code of PMBS is given in Alg.~\ref{alg:pmcts} with the selection subroutine given in Alg.~\ref{alg:pmcts-select}. Other subroutines of PMBS are mostly straightforward. 


\vspace{-3mm}
\begin{algorithm}
    \begin{small}
    \DontPrintSemicolon
    \SetKwFunction{FSelect}{Selection}
    \SetKwFunction{FExpand}{Expansion}
    \SetKwFunction{FRollout}{Rollout}
    \SetKwFunction{FBackpropagation}{Backpropagation}
    \Fn{\FSelect{$n_0$}}{
        $\text{Pairs} \gets \varnothing$ \;
        \While{\normalfont $len(\text{Pairs}) < N_\text{e}$}{
            $n^i \gets \text{traverse tree until a leaf node using Eq.~\ref{eq:ucb2}}$ \;
            $a^i \gets \text{one sampled action of } n^i$ \;
            Remove $a^i$ from the sampled actions of $n^i$ \;
            $\text{Pairs} \gets \text{Pairs} \cup \{(n^i, a^i)\}$ \;
            $\hat{N}(n^i) \gets \hat{N}(n^i) + 1$ \;
            increment virtual counts of ancestor nodes of $n^i$
        }
        \Return Pairs
    }
    \caption{\label{alg:pmcts-select}
    Selection with Virtual Loss)}
    \end{small}
\end{algorithm}
\vspace{-5mm}

%% file: texs/experiments.tex
We evaluated the proposed system (PMBS) in a physics simulator (Isaac Gym) and on a real robot on adversarial test cases.
In comparisons to baseline and ablation studies, we observe significant improvements using the GPU-based physics simulator together with parallel MCTS, which brings episodic decision making for real robots closer to real-time, i.e., a single complex decision is made in a few seconds.
All experiments are evaluated on a desktop with an Nvidia RTX 2080Ti GPU, an Intel i7-9700K CPU, and 32GB of memory.

\subsection{Simulation Studies}
In this work, the simulated environment is built with Isaac Gym~\cite{makoviychuk2021isaac}, consisting of a Universal Robot UR5e with a two-finger gripper Robotiq 2F-85, and an Intel RealSense D455 RGB-D camera overlooking a tabletop workspace as shown in Fig.~\ref{fig:overview}.
The robot is in position-control mode, push and grasp actions are controlling the position of the end-effector, inverse kinematic~\cite{hawkins2013analytic, feng2021team} is applied to convert to joint space.
The effective workspace is at a size of $0.288\times0.288$m, discretized as a grid of $144\times144$ cells where each cell is one pixel in the image (orthographic projection) taken by the camera.
%
The workspace, in comparison to previous studies~\cite{huang2021visual, huang2022interleaving}, is significantly smaller (only about $45\%$ in terms of area), making the setting much more challenging. 
We intentionally selected the setting to demonstrate the power of PMBS. 

\begin{figure}[ht!]
\vspace{-3mm}
    \centering
    \includegraphics[width=\linewidth]{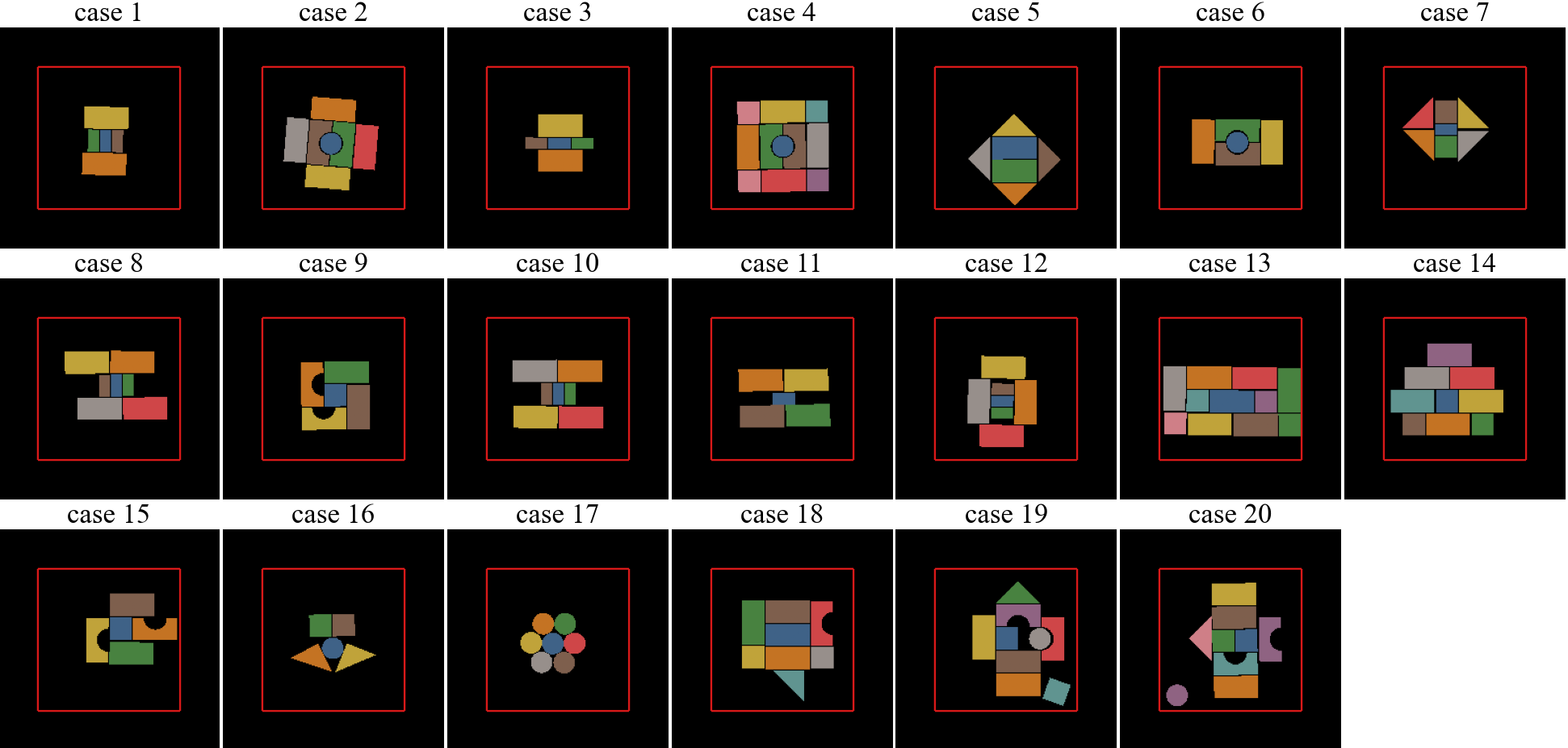}
    \caption{\label{fig:cases}
        20 cases~\cite{huang2021visual} used in simulation experiments, where the target object has a \textcolor{RoyalBlue}{blue} mask. No object should exceed the boundary (\textcolor{red}{red} lines).
    } 
\vspace{-4mm}    
\end{figure}

All objects should reside in the workspace. 
20 cases~\cite{huang2021visual} used for evaluation can be found in Fig.~\ref{fig:cases}, where the red lines denote the boundary to which objects centers must be confined at all times.
The push distance of a push action is fixed at $5$cm ($10$cm in previous work~\cite{huang2021visual, huang2022interleaving}), the effective push distance is around $3$cm (the distance objects are moved).

\noindent \textbf{Metrics.}
Four metrics are used to evaluate our systems: 
    1) the number of actions used to retrieve the target object,
    2) the total planning time used for retrieving the target object (build the tree),
    3) the completion rate in retrieving the target object within $16$ actions,
    4) the grasp success rate, which is the number of successful grasps divided by the total number of grasping attempts.

\noindent \textbf{Baseline.}
We use an optimized serial MCTS implementation as the baseline, where the number of environments used for MCTS is one.
The following hyperparameters are used across all methods in benchmark unless otherwise mentioned.
The discount factor $\gamma=0.8$.
The default max depth of tree $d_T=7$, and the default simulation depth $d_s=3$.
The threshold of GC is $R_c^{*} = 0.9$.
The UCB exploration term $c$ in Eq.~\ref{eq:ucb1} and~\ref{eq:ucb2} is $0.3$.
The time limit (budget) $T_{\text{max}}$ for one step planning is $60$ seconds.
$1000$ robots (environments) in Isaac Gym are used in our PMBS; it takes around $2.2$ seconds for all robots to complete one push action.

\begin{figure}[ht!]
\vspace{-3mm}    
    \centering
    \includegraphics[width = .97\linewidth]{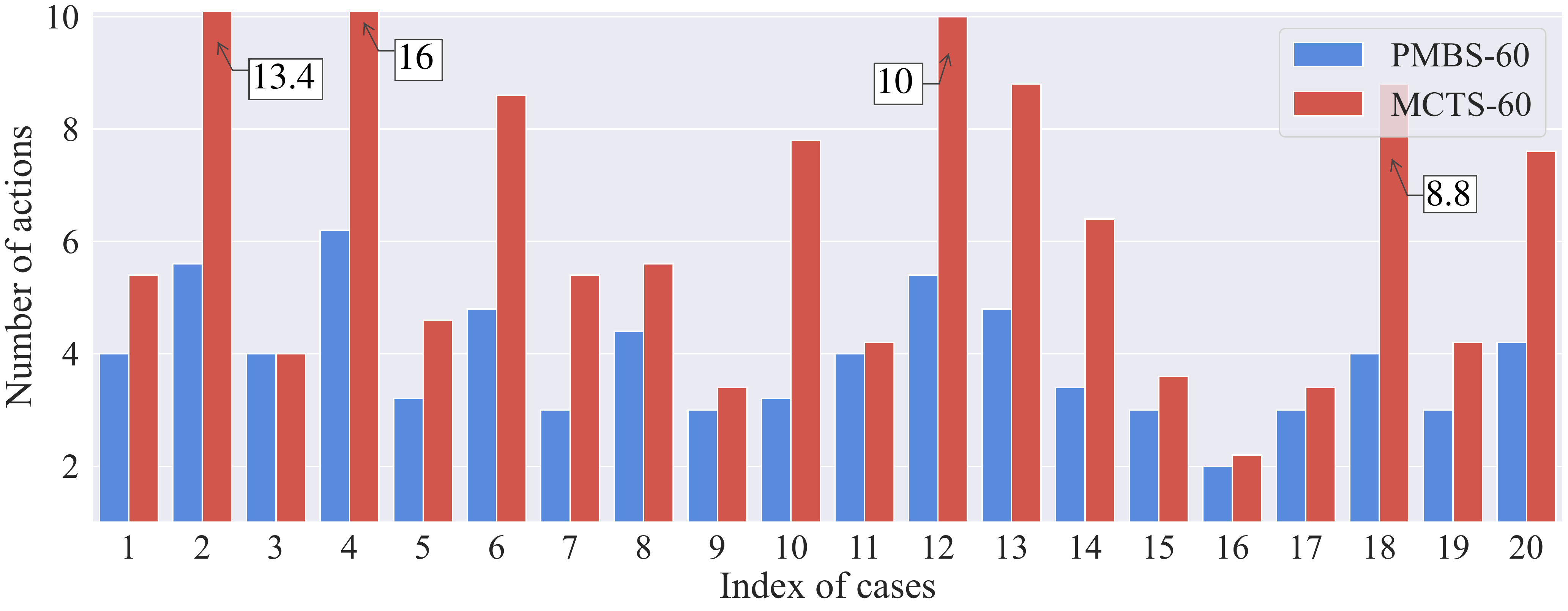}
\vspace{-2mm}    
    \caption{\label{fig:20-sim-num}
        The average number (over five independent trials) of actions per case needed for solving the twenty cases, given a time budget of 60 seconds.
    } 
\vspace{-2mm}    
\end{figure}

\begin{figure}[ht!]
\vspace{-4.5mm}    
    \centering
    \includegraphics[width = .97\linewidth]{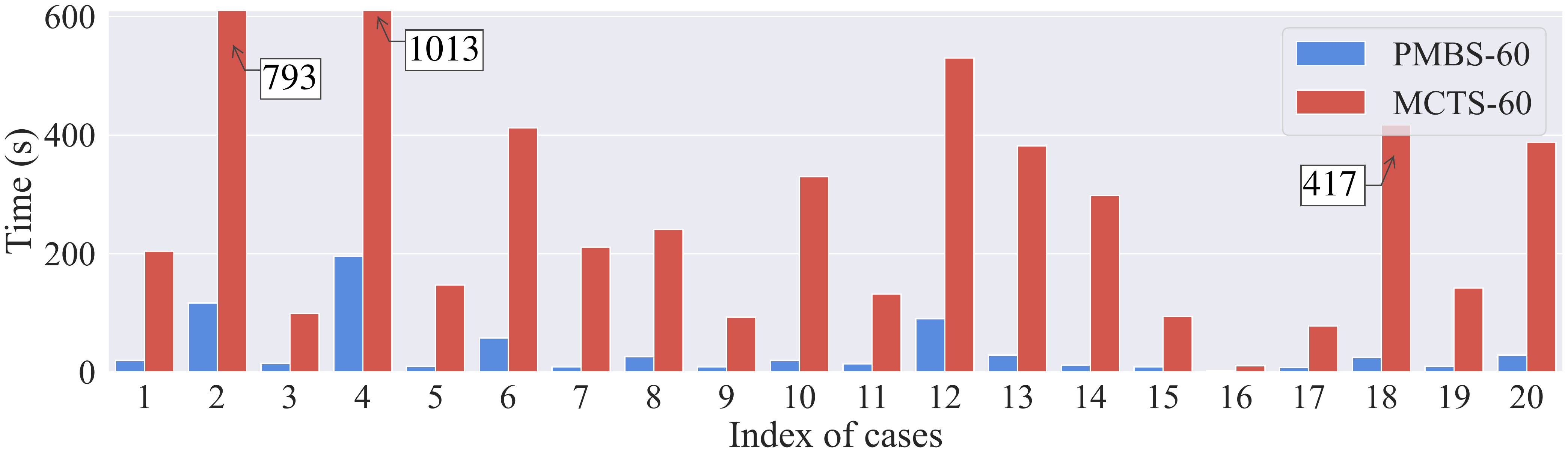}
\vspace{-2mm}    
    \caption{\label{fig:20-sim-time}
        The average time (over five independent trials) per case needed for solving the twenty cases, given a time budget of 60 seconds.
    } 
\vspace{-5mm}    
\end{figure}

We evaluate the performance of PMBS and the baseline serial MCTS over all 20 cases, running each case five times. For the evaluation, we set a time budget $T_{\text{max}} = 60$ and denoted the two methods as PMBS-60 and MCTS-60, respectively. 
The summary benchmark for these two methods can be found in the first two rows of Table.~\ref{tab:sim20table}; individual results for each case can be found in Figs.~\ref{fig:20-sim-num} and~\ref{fig:20-sim-time}.

\begin{figure}[ht!]
\vspace{-3mm}    
    \centering
    \includegraphics[width = \linewidth]{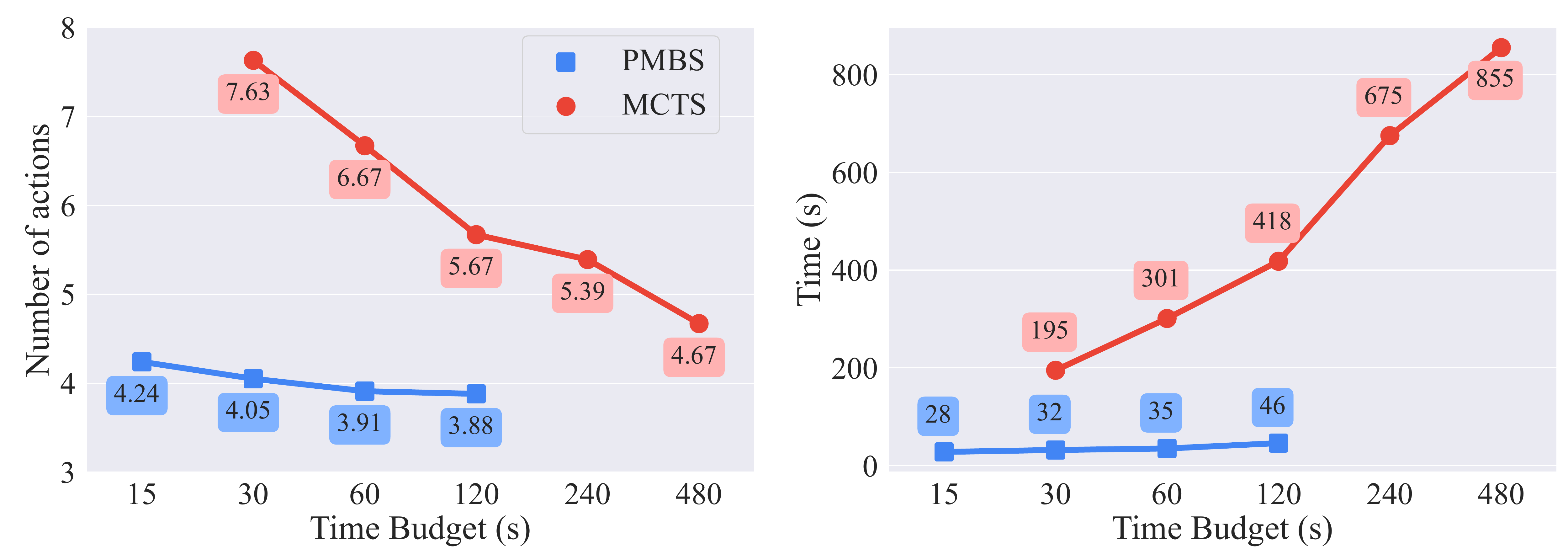}
    \caption{\label{fig:20-sim-ablation}
        PMBS and serial MCTS evaluated with different time budgets. The reported values are averages over all 20 cases.
    } 
\vspace{-4mm}    
\end{figure}

We make some observations based on the result. First, PMBS outperforms the serial MCTS version in terms of number of actions and computation time across all cases, which is as expected because PMBS engages many environments to facilitate its search effort. On the other hand, when we view the solution quality and computation time together, the advantage of PMBS over serial MCTS is significant: PMBS-60 uses 35 seconds on average for planning, whereas MCTS-60 uses over 300 seconds. This along translates to an $8.6\times$ speedup. At the same time, PMBS-60 uses $70\%$ fewer actions in solving the tasks. A further data point regarding the speedup at the same solution is given a bit later in Fig.~\ref{fig:20-sim-ablation}.

A second observation is that, despite the fact that we are dealing with a difficult long-horizon planning problem, PMBS is able to achieve planning that is close to being able to perform reasoning in real-time, as it takes an average of $35/3.91 < 9$ seconds to make a single decision. With further optimization and/or better hardware, we believe that PMBS will achieve real-time decision-making capability for the current set of object retrieval tasks. 

\begin{table}[ht!]
    \centering
    \scalebox{0.95}{\begin{tabular}{c|c|c|c|c}
        & Num. of Actions & Time & Completion & Grasp Success   \\ \hline
        PMBS-60 & $\mathbf{3.91}$ & $\mathbf{35}$s & $100\%$ & $98.3\%$ \\ \hline
        MCTS-60 & $6.67$ & $301$s & $93.0\%$ & $96.4\%$ \\ \hline
        PMBS-60 ($c = 0$) & $4.03$ & $113$s & $100.0\%$ & $99.2\%$ \\ \hline
        PMBS-60 ($c = \infty$) & $12.71$ & $147$s & $42.0\%$ & $96.7\%$ \\ \hline
    \end{tabular}}
    \caption{Simulation experiment results for 20 cases. Time budgets are limited up to 60 seconds.}
    \label{tab:sim20table}
\vspace{-2mm}    
\end{table}

\noindent \textbf{Ablation Study.}
The time budget $T_{\text{max}}$ is one of the main factors that influence the solution quality and planning time.
To understand its role, several time budgets are used to evaluate our method, as shown in Fig.~\ref{fig:20-sim-ablation}.
Given more time for tree search, serial MCTS and PMBS could improve the solution quality, leading to fewer required actions.
The trends of serial MCTS (number of action) are steep, as it is highly possible that it could not find a solution given a limited time. 
The trend of PMBS (planning time) is more gradual, as the most time-consuming search happens in the first few iterations, which usually uses all the time budget.
While serial MCTS never achieves the same solution quality as PMBS, comparing the first PMBS data point and the last serial MCTS data point, we observe a $855/28 = 30\times$ speedup with PMBS still has some quality advantage. 

On the flip side, we note that the speedup of $30\times$ seems small considering that we used $1000$ environments. This is due to two factors. First, MCTS is itself a serial process;  
parallelization will incur performance loss. Second, while we have improved many bottlenecks, e.g., on action sampling and grasp classification, the object retrieval task contains many elements that cannot be readily parallelized. 

We also evaluated the impact of the exploration and exploit trade-off on PMBS.
If the $c$ in Eq.~\eqref{eq:ucb2} is set to $0$, i.e., pure exploitation, PMBS-60 uses $4.03$ actions and $113$ seconds (planning time) in average on 20 cases.
The performance is worse than when $c=0.3$, as shown in Table.~\ref{tab:sim20table}.
This is expected as the greedy approach could be stuck in a local optimum.
PMBS-60 is also tested by setting $c$ to be a large number in Eq.~\eqref{eq:ucb2}, i.e., pure exploration, which uses $12.71$ actions and $147$ seconds (planning time) on averages; the completion rate has a steep drop to $42.0\%$.

\subsection{Real Robot Experiments}
\begin{figure}[ht!]
\vspace{-2mm}   
    \centering
    \includegraphics[width = \linewidth]{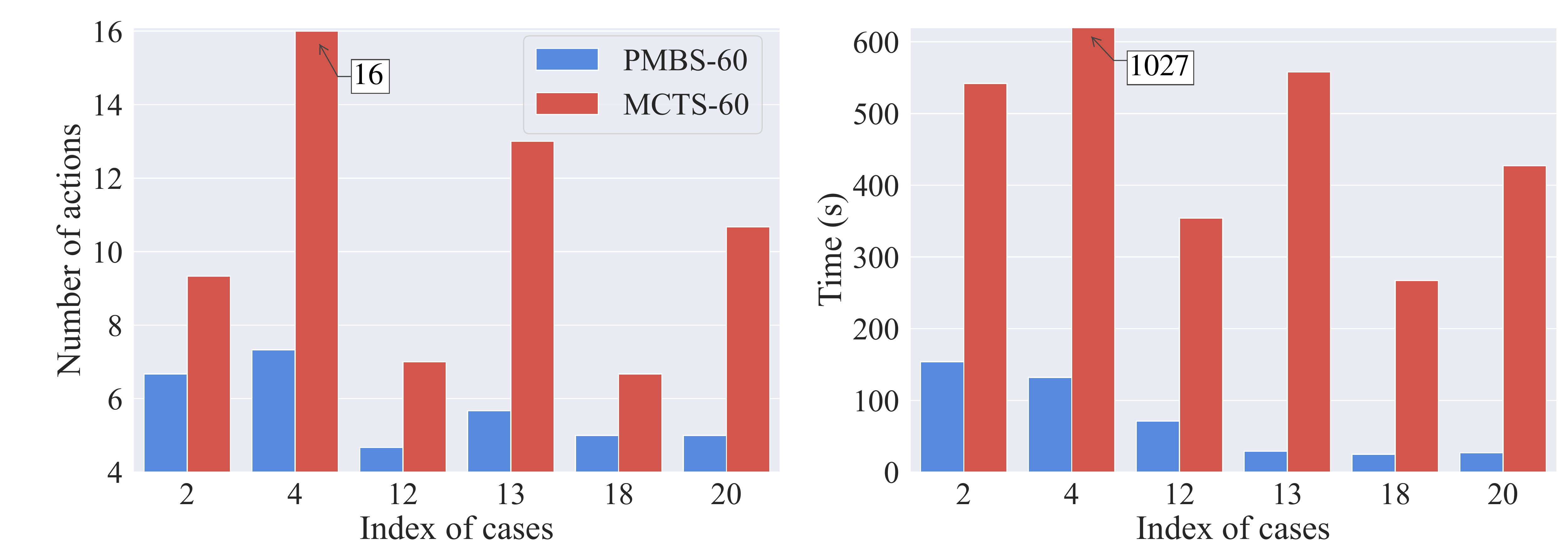}
    \caption{\label{fig:real-num-time}
        The number of actions and time used for solving the six most challenging cases on the physical robot.
        The time budget is $60$ seconds.
    } 
\vspace{-2mm}    
\end{figure}

For experiments on the physical UR-5e, the input to PMBS is a single RGB-D image.
A $1280\times720$ RGB-D image is taken, then it is orthogonally projected over the workspace of resolution of $144\times144$. 
Since the same robot and objects are used in both simulation and the real world, we can observe and act on a real robot but plan in a simulator.
\changed{
For each object, simple pose estimation is performed to load objects from real images to the physics simulator environments.
}
\changed{The pose estimation is done by firstly extracting masks for objects from the image, then a brute-force matching between detected mask and recorded mask is performed for each object. We could achieve it at 0.15 seconds for one image (around 10 objects).}
Serial MCTS and PMBS are evaluated the same way as in simulation experiments, except we only run tests on the six most challenging cases.
Individual benchmark on six cases can be found in Fig.~\ref{fig:real-num-time}.
Average statistics are listed in Table.~\ref{tab:real6table}.
We observe minimal sim-to-real performance loss; 
A small gap exists between the real and the simulation experiments, mainly due to pose estimation errors and mismatch of physics properties.

\begin{table}[ht!]
\vspace{-2mm} 
    \centering
    \begin{tabular}{c|c|c|c|c}
        & Num. of Actions & Time & Completion & Grasp Success   \\ \hline
        PMBS-60 & $\mathbf{5.72}$ & $\mathbf{73}$s & $100\%$ & $100\%$ \\ \hline
        MCTS-60 & $10.45$ & $529$s & $83.3\%$ & $87.0\%$ \\ \hline
        PMBS-60 (sim) & $5.03$ & $81$s & $100\%$ & $97.2\%$ \\ \hline
        MCTS-60 (sim) & $10.77$ & $587$s & $76.7\%$ & $96.6\%$ \\ \hline
    \end{tabular}
    \caption{Real robot experiment results on the six most difficult cases. Time budgets are limited to 60 seconds per case.}
    \label{tab:real6table}
\vspace{-2mm}    
\end{table}

\textbf{Additional Experimental Details.} Curious readers may find in the online supplementary material additional experimental details including complete, actual execution snapshots of PMBS and MCTS for all 20 cases, as well as the execution snapshots for real robot experiments.

%% file: texs/conclusion.tex
In this work, we propose PMBS, a novel parallel Monte Carlo tree search technique with GPU-enabled batched simulations for accelerating long-horizon, episodic robotic planning tasks. Through carefully making a series of design choices to overcome multiple major bottlenecks resisting the parallelization effort, PMBS achieves over $30\times$ speedups compared to a decent serial MCTS implementation while still delivering better solution quality, using the same computing hardware. Real robot experiments show that PMBS directly transfers from simulation to the real physical world to achieve near real-time planning performance in solving complex long-horizon episodic robot planning tasks. 